# An Efficient Pattern Mining Convolution Neural Network (CNN) algorithm with Grey Wolf Optimization (GWO)

*1Aatif Jamshed*, 2Dr. Bhawna Mallick, 3Dr. Rajendra Kumar Bharti,

*1Research scholar, Department of Computer Science and Engineering, Uttarakhand Technical University, Dehradun, Uttarakhand, India.

2Senior consultant at Maverick Quality Advisory services Pvt, Ltd, Uttar Pradesh, India.

3Associate Professor, Bipin Tripathi Kumaon Institute of Technology, Dwarahat, Uttarakhand, India.

*Corresponding author email: aatifjamshed@yahoo.com/09.aatif@gmail.com

**Abstract**

Automation of feature analysis in the dynamic image frame dataset deals with complexity of intensity mapping with normal and abnormal class. The threshold-based data clustering and feature analysis requires iterative model to learn the component of image frame in multi-pattern for different image frame data type. This paper proposed a novel model of feature analysis method with the CNN based on Convoluted Pattern of Wavelet Transform (CPWT) feature vectors that are optimized by Grey Wolf Optimization (GWO) algorithm. Initially, the image frame gets normalized by applying median filter to the image frame that reduce the noise and apply smoothening on it. From that, the edge information represents the boundary region of bright spot in the image frame. Neural network-based image frame classification performs repeated learning of the feature with minimum training of dataset to cluster the image frame pixels. Features of the filtered image frame was analyzed in different pattern of feature extraction model based on the convoluted model of wavelet transformation method. These features represent the different class of image frame in spatial and textural pattern of it. Convolutional Neural Network (CNN) classifier supports to analyze the features and classify the action label for the image frame dataset. This process enhances the classification with minimum number of training dataset. The performance of this proposed method can be validated by comparing with traditional state-of-art methods.

## 1. INTRODUCTION

IN the recent development process of image processing system, most of the applications were focused on the various kinds of real time image frame classification system with enhanced prediction model of human action recognition. This improves the automation process which is to reduce the error rate in the classification system of image frame validation [1]. To classify or predict the type and human action in a video / image frame, there are several sources of instruments and devices to get the video in the range of low resolution and also in the high-resolution factor. From that video data, the frames were split to form image frames and by the image processing techniques leads to predict and analyze the captured picture by segmenting and classifying the activity of human in the video. The segmentation and classification process are deals with the image pixel analysis based on certain rules and conditions [2]. This is to validate the pattern of image where the region of interest (ROI) present in the image frame. There are several methods to identify the location of ROI in the image which are based on the pattern extraction and clustering the particles that are related to the required object [3]. In that, the correlation factor among the current frame and predefined features were classified to get the ROI.

There are several applications in the video processing to analyze the source of video such as moving object tracking, vehicle tracking and recognition, gait recognition, etc. that are based on the image geometrical structure [4] and pattern-based image analysis method to classify the required object with high accuracy compare to other traditional feature analysis methods. Since, the clustering model for ROI segmentation refers the relevance pixel intensity value for the overall image matrix and also consider the histogram peaks of image [5]. The segmentation and classification both we integrated together to form the better image analysis model. For the enhanced classification performance, the Convolutional Neural Network (CNN) [6] which can extract the features based on the block separation.

In the traditional model of image classification, the feature vector we need to extract from image matrix by using some feature extraction methods. In CNN, the convolution pattern of an image was considered as the feature vector and can be classified based on the neuron network formation while at the training stage of CNN classifier [7]. This type of classification process doesn't require any type of feature extraction method to represent the image feature set. Since, this CNN model evaluates the pattern extraction model internally for the block separated image.

From the previous discussion, the image pattern extraction method enhanced the recognition rate of an image processing application. In that, the CNN was also considering as the pattern-based image recognition system that is to recognize the image based on the convolution pattern of image [8]. To improve the performance of CNN, the image texture pattern was used as the input of classifier instead of the direct input of image data. There are several types of texture pattern extraction methods have been implemented for the image / video classification and human action prediction process such as Local Binary Pattern (LBP), Local Tetra Pattern (LTrP), Local Ternary Pattern (LTP), and other pattern extraction methods [9]. In this proposed model of human action recognition system, a novel image texture pattern extraction method was implemented to predict the difference of neighborhood pixels in magnitude in different angles of projection model. This can be achieved by using Convoluted Pattern of Wavelet Transform (CPWT) texture extraction method with the combination of wavelet transform. In this the wavelet transform is for reducing the dimensionality of image feature and the convolution pattern is the result of convoluted image with pattern analysis. These combinations enhanced the pattern retrieval model and improve the classification performance of CNN. Further the feature can be optimized by using the Grey Wolf Optimization (GWO) method [10] that selects the best feature attribute among the overall feature set. According to the selected attributes, the neural network in the CNN was formed while at the training process of that classifier. The motivation of this proposed work is to extract the better pattern extraction model for the video processing and reduce the complexity of it. In that, the frame pattern from wavelet transform provides better feature model. Along with that GWO optimization improves the feature learning by optimally selects the feature attributes for classification. To achieve this, the Convoluted Pattern of Wavelet Transform (CPWT) and GWO optimization are proposed. The main motivation to implement the CNN classifier is that it can recognize the features of texture pattern better than other existing classification model. The novelty of this paper is to extract the features for the frame based on the texture pattern and optimal feature learning by using the combination of CPWT with GWO based pattern analysis. The CPWT-GWO enhances the pattern leaning model for CNN and to improve the classification efficiency.

The main contribution of the proposed work is

1. To equalize the image intensity of the image and to normalize pattern, the Cellular Automata based image filtering method.
2. A novel method of image pattern extraction by using CPWT method.
3. To classify the category of human actions in a video frame by using the CNN classifier.
4. To reduce the feature size and to select best attribute by using the GWO optimization method.

The paper work have been segmented into following sections: The survey of various methods of feature extraction, feature selection and classification algorithms was presented in section II. Then the overall proposed work and its algorithm procedure were described in section III. The section IV presented the analysis of different pattern extraction methods that are compared to the proposed model of recognition system based on the CPWT pattern extraction and GWO optimization with CNN model. The results are justified and concluded with future enhancement in section V.

## 2. RELATED WORK

This section surveys the various types of image frame feature extraction methods and to classify the image frame categories. The major topics that are discussed in the survey is based on the texture pattern extraction models and classification methods in the endoscopic and in other image frame processing applications.

In [11], author proposed a dynamic feature learning system based on the skeleton structure of video sequence. In that, the spectral of body parts were extracted and classified by using the CNN method. In [12], paper discussed about the first-person action recognition model to classify the video appearance, shape and motion of the visual rhythm. This was extracted by referring the textural features of the video dataset. From this type of texture analysis, [13] proposed texture fusion based facial expression recognition model. In that, the LBP and landmark based Active Shape Model (ASM) was fused together to form the texture plus geometrical feature analysis model which is classified by the Support Vector Machine (SVM) method. In [14], author proposed the action recognition model from RGB with Depth video dataset. To extract the RGB and depth image / video data, the Kinect sensor-based video system was used. To classify the human action, deep learning method was implemented with extreme learning machines. Later, to enhance the prediction performance for the RGB-D image dataset, [15] proposed multi-directional projected depth motion mapping technique was implemented and tested for the action recognition system. This was further improved by embedding the skeleton key points of the human body [16 & 17]. In [18], the texture pattern was extracted to map the magnitude and parameters representation. The Depth image sequence became an important feature for the action prediction system. By considering that, paper [19] proposed a R-pattern transformation method combined with Zernike Moments to recognition the human abnormal action for the depth image sequence of video data. Like that the same, [20] proposed a fusion based machine learning model based on the SVM classifier and the Nave Bayes classification model to recognize the action for both RGB and the depth image. In [21], the skeleton structure-based action prediction by using the hierarchical spatial feature learning combined with temporal stack learning network architecture. The human action recognition was further enhanced by the segmentation approach based on statistical weight of image pattern in [22]. This was optimally selected by the rank correlation method.

In the image classification process, most of the applications were developed by using the convolutional neural network to improve the accuracy of prediction. According to that, [23] proposed a salient feature analysis based on the 3D CNN model with LSTM method. This can reduce the size of overall feature by considering the salient features of the video. In [24], author proposed a novel model of Deep CNN based action prediction model for the still images. In this, the CNN was fused with two or three number of concatenated models. To enhance the feature learning of CNN, [25] proposed an efficient image pooling model of action recognition system. This image pooling method reduce the dataset of video frames instead of analyzing the whole video sequences. Also, [26] proposed a hint enhanced deep learning method for action recognition. In this paper, the investigation is based on the potentials of CNN for the image-based action recognition. This type of CNN based action recognition system was implemented in the 3D skeleton structure of human video based on the scale invariant mapping and the multi-scale dilated CNN technique in [27]. Similarly, the paper work [28] proposed two-stream based action recognition using CNN. For the LSTM based CNN, [29] proposed a quanternion spatial temporal model was developed for the RGB video action recognition system. Compare to the traditional CNN model, the QST-CNN enhance the feature learning concept of CNN. In [30], the paper proposed the review of different techniques of CNN model for the action recognition-based application. For the RGB-Depth image classification, the CNN extracts the textural feature to analyze the rotational invariant vector. In [31], author proposed depth motion maps based Local Ternary Pattern based texture pattern extraction method to combine the spatial feature and textural feature of the image sequence. For the cloud-based application, [32] proposed super-pixel transformation-based feature retrieval model to improve the efficiency and increase the speed of performance compare to tradition model of CNN. In tis the human action recognition is processed by four stream deep convolutional neural network method. To reduce the feature size for learning process of CNN classifier, the image feature was optimized by using the Artificial Bee Colony (ABC) optimization algorithm [33]. This can select the best feature attributes among the overall feature set. Similarly, in [34], author implemented the optimal feature selection by using the combination Gray Wolf Optimization (GWO) algorithm with the CNN classifier. In [35], author proposed hierarchical spatial-temporal dependent feature analysis model with CNN to enhance the classification performance of CNN with LSTM model.

In [36], the paper surveyed different fusion technique of depth image with RGB to recognize the human action. The multi-class SVM and other classification type of neural network were presented for the action recognition. This was then improved by using the hybrid method of heuristic algorithm with deep learning model in [37]. In this, the ABC optimization and the PSO are combined to form the hybrid optimization technique for feature selection. This improves the classification performance. A survey of vision based human action recognition is presented in [38]. In this survey, the problems facing in the vision capturing was focused such as camera motion, dynamic background and other impacts that are failed in the previous methods of action recognition. From this analysis, the paper [39] presented the action recognition by using the LSTM method. In this, the LSTM is improved by two-stream attention model. This improves the learning structure of the classifier better than the other traditional model. In [40], author proposed a two-fold transformation model with the Gabor Wavelet Transform (GWT) and Ridgelet Transform (RT). In this, the human action recognition using decisive pose.

From these review of existing work for the image frame feature analysis, most of the author proposed in the texture pattern-based feature extraction methods and also justifies with better enhancement over another method to improve the classification performance. The survey collections consist of both action prediction and other image frame processing applications to evaluate the performance of pattern extraction method in various domains. From this survey, its clears that in pattern extraction method, the multi-angular based pixel validation achieved better accuracy than those methods that are referring the smaller number of boundary validation. Since, in these methods, the spatial representation of the image frame helps to improve the recognition rate with clear depth of pixel intensity. This form of spatial image frame representation and multi-angular pixel validation can be achieved by the proposed method of Convoluted Pattern of Wavelet Transform (CPWT)compared with Local Binary Pattern (LBP). The detailed description with algorithm steps and the result validation of proposed method were discussed in the following sections.

3.  PROPOSED WORK

A novel model of image pattern recognition model was proposed in this paper with the CNN classification. In this proposed feature analysis, the feature vector from Convoluted Pattern of Wavelet Transform (CPWT) for the testing image frame in a video was optimally selected by using the Grey Wolf Optimization (GWO) algorithm. This combination performs a multi-angular pattern analysis model to improve the performance of classification. In this the texture pattern from the CPWT was consider as the input for CNN in both training and testing feature set. This type of feature pattern improves the network formation in the CNN to identify the relevancy between pattern from training and the testing feature. In this, the GWO act as the dimensionality reduction by referring the histogram peaks of image pattern.

The important key points in the proposed pattern analysis model can be listed as,

1. The pre-processing can be processed by using the Cellular Automata based filtering method to reduce the noise in the image pixel and enhanced the edge information to extract pattern.
2. The texture pattern of the image frame can be extract by using the convolution pattern integrated with wavelet transform of CPWT method.
3. The pattern features can be optimally selected as the best texture for classification is by using GWO optimization algorithm.
4. Then these texture features of an image frame can be classifying by using the CNN method to enhance the prediction performance.

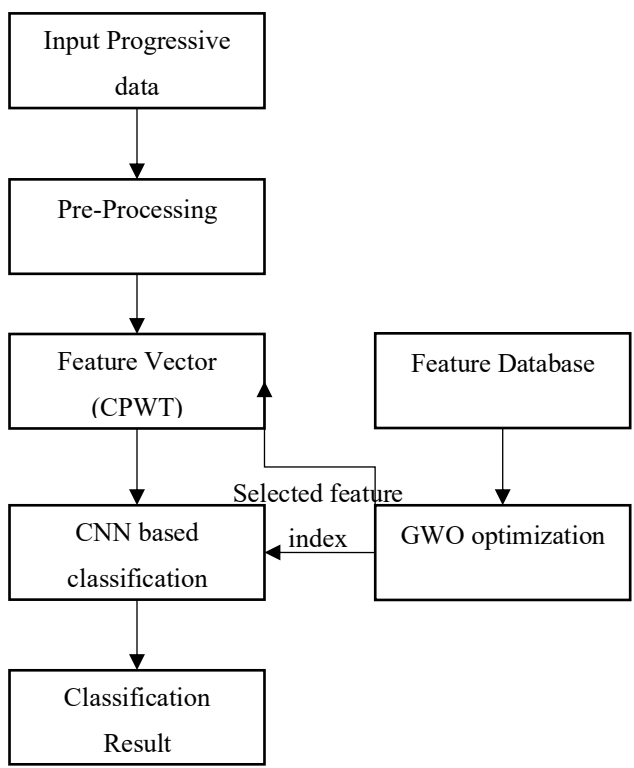

**Figure 1:** Overall flow diagram of proposed image analysis model

The overall block diagram of the proposed feature analysis model is shown in the figure 1 with line flow for the application of video retrieval system. In that, the pattern of the image can be extract by verifying the difference in pixel intensity between the neighboring pixels. Here, the mask of the image can choose based on the number of neighboring pixels that are selected from the mask of that image. Then from that encoded image, wavelet transform is used to get the full texture of CPWT. In that, the GWO optimization is used to select the best feature of that image by referring the histogram peaks of that pattern. From the histogram peaks, the best region was selected to represent the feature of an image frame. This was processed for the frames in a video of source. Finally, the average rate of classification presents the recognized result class to indicate the action of the video.

The functions that are included in the proposed model can be listed as,

a) Pre-processing (CA),
b) Texture Pattern Extraction (CPWT),
c) Feature Optimization (GWO)
d) Texture Classification (CNN).

## A. Pre-processing

In image processing, the image noise filtering and the image equalization techniques were most commonly used in the pre-processing stage. This is to enhance the pixel quality by normalizing the pixel intensity to make the equalization of histogram in an image. There are several types of traditional filtering algorithms were implemented to remove the noise by applying smoothening effect to the image. This can be processed by using the image transformation methods. In that some of the methods were approaching the concept of neighborhood pixel validation. The methods like median filter, mean filter and other types of statistical properties improves the noise removal rate better than other methods. In related to that, the Cellular Automata (CA) also refers the neighborhood pixels to filter the image. These types of filters used some standard size of mask to apply over the image for noise identification and normalization.

The figure 2 shows the sample matrix with indexing labelled mask that are generally used in the filtering model. In that, the mask size was selected in the basis of odd value which can be as like 3×3 and 5×5 matrix. While choosing the mask size, the smaller size needs to consider for the better Peak Signal to Noise Ratio (PSNR) value and lower the Mean Square Error (MSE). By considering this, the proposed filtering technique implements 3×3 mask size. In that, the detected noisy pixel forms the image can be normalizing by applying the Laplacian transform over the separated mask of the image frame. The detailed steps that are involved in the CA filtering with Laplacian transform is stated in the algorithm 1.

**Algorithm 1:** CA filtering

**Input:** Testing Video input, $V$

**Output:** Filtered video frame, $V_f$

Convert video '$V$' to image frames '$I_{in}$'

Laplacian transform for image frame by equation (1).

**for** m = 2 to R-1 **do**

**for** n = 2 to C-1 **do**   // 'R' and 'C' are row and column of image frame size.

$I_X = I_L(m-1:m+1, n-1:n+1)$     // Separating image to cells.

Estimate differences in $I_b$ and $I_c$. // $I_b$ - Boundary Pixels, $I_c$ - Center of mask.

Estimate magnitude of the difference $I_p$ by equation (3).

**If** $I_b \sim I_c > I_p$, **then**

$t = I_p$

**Else**

$t = I_c$

**End if**

$I_f(m,n) = t$  // Filtered image frame output

**End 'm' loop**

**End 'n' loop**

---

The overall block diagram of the proposed CA with Laplacian filtering model is shown in the figure 2.

The Laplacian transform can be estimate for the image frame '$I_{in}$' which is represent as '$I_L$'. This can be calculating by the equation (1).

$$I_L = \sum (I_{in} * M) \qquad (1)$$

Where, 'M' – Mask matrix of Laplacian distribution based filter.

Here the error pixel can be estimate by the equation (2).

$$E_{xy} = \begin{cases} P_{mn}, & if(N_{min} \geq P_{mn} \geq N_{max}) \\ 0, & Otherwise \end{cases} \qquad (2)$$

Where, $P_{mn}$ – Noisy pixel

N – Number of boundaries.

| m-1, n-1 | m, n-1 | m+1, n-1 |
|---|---|---|
| m-1, n | m, n | m+1, n |
| m-1, n+1 | m, n+1 | m+1, n+1 |

(a) 3×3

| m-2, n-2 | m-1, n-2 | m, n-2 | m+1, n-2 | m+2, n-2 |
|---|---|---|---|---|
| m-2, n-1 | m-1, n-1 | m, n-1 | m+1, n-1 | m+2, n-1 |
| m-2, n | m-1, n | m, n | m+1, n | m+2, n |
| m-2, n+1 | m-1, n+1 | m, n+1 | m+1, n+1 | m+2, n+1 |
| m-2, n+2 | m-1, n+2 | m, n+2 | m+1, n+2 | m+2, n+2 |

(b) 5×5

**Figure 2** Schematic structure of CA mask

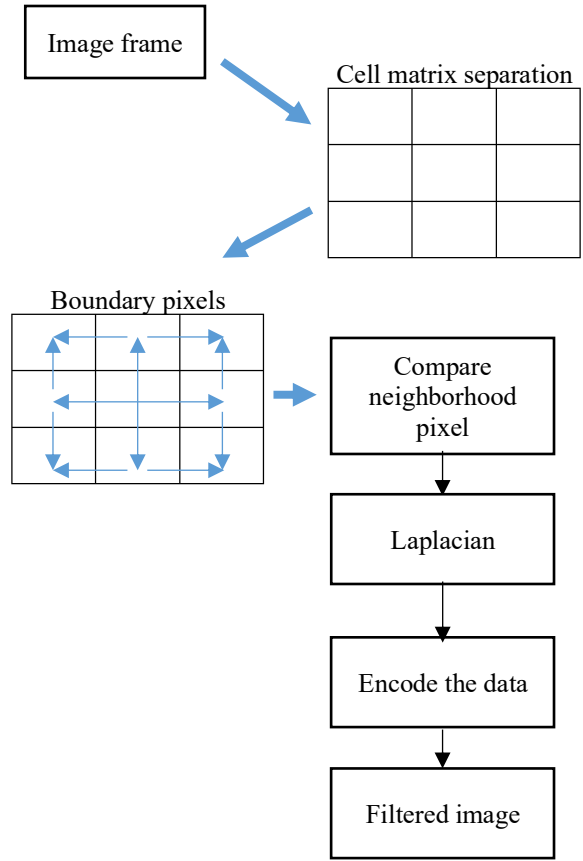

**Figure 3** Block diagram of CA-Laplacian

The predicted noisy pixel and the position of it can be normalize by estimating the magnitude of difference between the neighboring pixels. This was represented as $I_b$ and can be evaluate by the equation

$$I_b = \frac{\sum_{k=1,3}^{L-1}(I_X)}{L} \quad (3)$$

Where, L – Length of boundaries in the image frame mask.

### B. Texture Pattern Extraction (CPWT)

The detailed description of the proposed texture pattern model of CPWT is presented in this section. Here, the combination of wavelet transforms and proposed model of CPWT algorithm is to represent the texture of each image frame from the pre-processed video input. The algorithm 2 describes the steps in the CPWT pattern extraction method.

---

**Algorithm 2:** CPWT algorithm for pattern extraction

**Input:** Filtered Image frame $I_f$.
**Output:** Image frame Pattern output $I_p$
Initialize Convolution Mask, $G_M$
In this, 'M' –matrix in the size of 5×5 for the image frame pattern.
**For** $m = 3\ to\ M - 2$     // Loop for row index 3 to M-2 size
**For** $n = 3\ to\ N - 2$ // Loop for column index 3 to N-2 size
$I_K = I_f(m - 2: m + 2, n - 2: n + 2)$  // Image frame into 5×5 matrix of cells based on the 'm' and 'n' index.
The cell separated patch can be convolute by equation (4),
The image gradient for the convoluted frame can be estimate by equation (5),
For various projection angles $\{-45^0, 0^0, +45^0, +90^0\}$, the magnitude was estimated by equation (6),
The maximum pixel progression value from the function for each pixel can be calculated by equation (7),

The iteration of 'k' value was varied to get the first and second maximum of the function as by iterating k = {1, 2}.
Encode the CPWT particle by equation (8), $I_p(m,n) = CPWT(x,y)$
**End For 'j'**
**End For 'i'**
Estimate Wavelet Transform using (9)
The histogram of $I_p$ is to represent the feature vector.

From the algorithm, the '$I_K$' represent the 5×5 splitted cell from the frames of input video. The '$C_I$' represent the convoluted result of the image frame. The equation (4) represents the convolution with mask matrix.

$$C_I = I_K * C_M \quad (4)$$

Where, $C_M$ – Convolution Mask matrix.

The '$|G_{xy}|$' and '$\alpha(x,y)$' represents the magnitude and gradient of the convoluted image matrix. These can be calculated by using the equation (5).

$$|G_{xy}| = \sqrt{G_x^2 + G_y^2}, \; \alpha(x,y) = \tan^{-1}\left(\frac{G_x}{G_y}\right) \quad (5)$$

Where, 'x' and 'y' are the mask cell size

$G_x = \frac{\partial C_I}{\partial x}$ and $G_y = \frac{\partial C_I}{\partial y}$

After getting the gradient and magnitude of image frame, then this can be quantized in different projection angles like $\{-45^0, 0^0, +45^0, +90^0\}$. This form of normalization can be estimate by using the equation (6).

$$I_{\alpha_L}(x,y) = \sum_{x=-N_1}^{N_1} \sum_{y=-N_2}^{N_2} |G_{xy}(x,y)| \times f_2(\alpha_L, \alpha_U, \alpha(x,y)) \quad (6)$$

Where, $f_2(p,q,r) = \begin{cases} 1 & if \; p \leq q < r \\ 0 & else \end{cases}$

$$\alpha_L = \{-45^0, 0^0, +45^0, +90^0\}, \alpha_U = \alpha_L - 45^0$$

The maximum pixel progression for each exported pixel of the pattern can be estimated by the equation (7).

$$\gamma_k(x,y) = \max_{\alpha_L}\left(I_{\alpha_L}(x,y)\right) \quad (7)$$

Finally, the CPWT pattern can be encoded from the binary representation of magnitude difference can be calculate by using the equation (8).

$$CPWT(x,y) = \sum_{i=0}^{P} 2^i \times f_3\left(I_{Ref}(s,t), I_{\gamma_1}(s,t), I_{\gamma_2}(s,t)\right) \quad (8)$$

Where, $I_{Ref}(s,t) = C_I(x+t, y+t), \quad \forall t = -1:1$

$$f_3(p,q,r) = \begin{cases} 1, & if \; p < r \& q \; OR \; p > r \& p \\ 0, & else \end{cases}$$

$$\begin{cases} \gamma_k = 1 \rightarrow m = 2; n = 0 \\ \gamma_k = 2 \rightarrow m = 2; n = 2 \end{cases}$$

The pattern extraction by the three function was related to the window size that are used in the cell separated filtering technique. The example of 5×5 size of image mask with its indexing are shown in figure 2(b). From the window matrix, the pixel magnitude was estimated in different projection angles as in the directions of $\{-45^0, 0^0, +45^0, +90^0\}$.

Finally, the wavelet transform for the extracted pattern can be represent in (9).

$$W(n) = \sum_{k=-\infty}^{\infty} X(k)h(2n-k) \quad (9)$$

The histogram of the wavelet pattern was represented as the feature vector from CPWT texture pattern.

C. *Feature Optimization (GWO)*

The Grey Wolf Optimization (GWO) algorithm solves the problem as like the way of hunting the prey by grey wolves.

Basically, there are four different categories of grey wolves such as Alpha (α), Beta (β), Delta (δ) and Omega (ω). In that the alpha type is considered as the head of that group, the beta type act as the decision maker, the delta type act as the food provider and the omega is act as hunting the prey. Likewise, the GWO arrange the searching and optimization structure to select the best attributes among overall feature set. The overall concept of GWO is referred from [24].

At first, the wolf particles are initialized with the size of feature attributes which can be represent as '$X_i$'.

(i.e.) $X_i = \{1,2, \ldots, n\}$

Then, the coefficient vectors 'a', 'A', and 'C' were initialized as the random value for indicating the parameters of current position.

At the startup, the fitness value of each search agent can be calculating for the 'α', 'β', and 'δ'. Now the position of these search agent can be estimate by the random value of vector 'A' and based on the velocity of particle movement.

Then this was iteratively updated in a loop until it reaches the maximum number of iteration count. The search agents update the position and find the prey near to that location and whether it is near to the radius of searching by updating position and velocity for each iteration. Once, if they found the prey in that radius limit, the omega group start to hunt the best one and this will be considering as the best solution. Then it searches for the nearest prey within the radius by estimating the fitness value.

From the paper work, the best features can be selected in the basis of validating the 'α', 'β', and 'δ' by updating the position and velocity at each iteration to find the best selection of attributes from histogram peaks.

### D. *Texture Classification (CNN)*

The CNN classifier was introduced to reduce the classification error rate due to different types of feature extraction techniques that are less robustness than compare to the CNN method. The CNN and its detailed algorithm procedure were described in [25]. In this, at the training stage of the classifier, the layers are to form the network between neurons that are based on the relevancy features in the dataset. During testing case, this estimates the distance between the training set and testing feature vector to retrieve the class label of testing image frame.

In the proposed work, the optimized feature was selected by the GWO and these are classified by estimating the probability of hypothesis between training and testing data. There are three major layers in the CNN that are can be listed as

1. Convolutional layer,
2. Max-Pooling layer, and
3. Fully-Connected layer.

In the training stage of CNN, these layers were mainly involved in the two major categories of propagations such as,

1. Forward propagation, and
2. Backward propagation.

In these two propagations, the one is for the evaluation of feature relevancy and another one is for validating the distance whether it is within the range or not. In that, the forward propagation contains the mathematical model to represent the network arrangement based on the feature matrix in the input and hidden layer of NN. In these 5 layers are used in the CNN to train the features and predict the relevancy among features. This was considered for the training case of dataset for the input patterns of image frame. Then during testing process, the feature vector was validated from the convoluted pattern and predict the class by identifying the minimum distance among the training feature set and the testing vector. The fully-connected layer present at the output stage of CNN that represents the classified result of it.

The convolution of image frame '$H$' can be evaluated by the equation (10).

$$H = X * f \qquad (10)$$

Where, '$X$' – Input image frame,

'$f$' – Filter mask for convolution.

In the CNN, the convoluted image frames were rearranged into several pre-defined blocks to make it into small patches. The size of patch separation is depending on the feature length that are to be classify. From these split blocks of image frame, the feature vector can be extract by estimating the maximum value of pixels in each block which are collectively arranged as the feature vector. This is pooling into the stream of data and passed into the input layer of neural network stage.

In this, the selection of forward and backward propagation can be select by referring the layers that are propagating through it. From that the result from CNN are arranged in the fully connected layer. In this, the convolution of the image frame from CNN can be defined as in equation (11).

$$y_{ij} = \sigma(x_{ij}) \qquad (11)$$

Where, $y_{ij}$ – Convolutional layer output by non-linearity of data.

$\sigma(x_{ij})$ – Convolution function.

The pre-nonlinearity of the data can be computed for $(x_{ij})$ can be written as

$$x_{ij} = \sum_{a=0}^{m-1} \sum_{b=0}^{m-1} \omega_{ab} y_{(i+a)(i+b)} \qquad (12)$$

Where, $\omega$ – Filter matrix.

The gradient error function $\left(\frac{\partial E}{\partial y_{ij}}\right)$ for each neuron output represents the convolution layer of back propagation. This can be represented as in equation (13).

$$\frac{\partial E}{\partial \omega_{ab}} = \sum_{i=0}^{N-m} \sum_{j=0}^{N-m} \frac{\partial E}{\partial x_{ij}} y_{(i+a)(i+b)} \qquad (13)$$

The basic block diagram of the CNN with both forward and backward propagation arrangement is shown in figure 4. The architecture diagram of CNN model is shown in the figure 5.

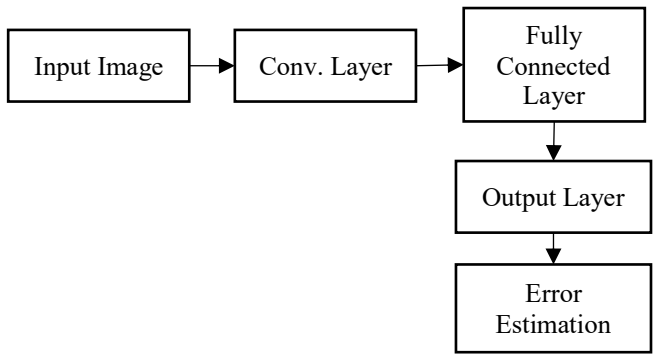

**Figure 4:** Basic block diagram of CNN

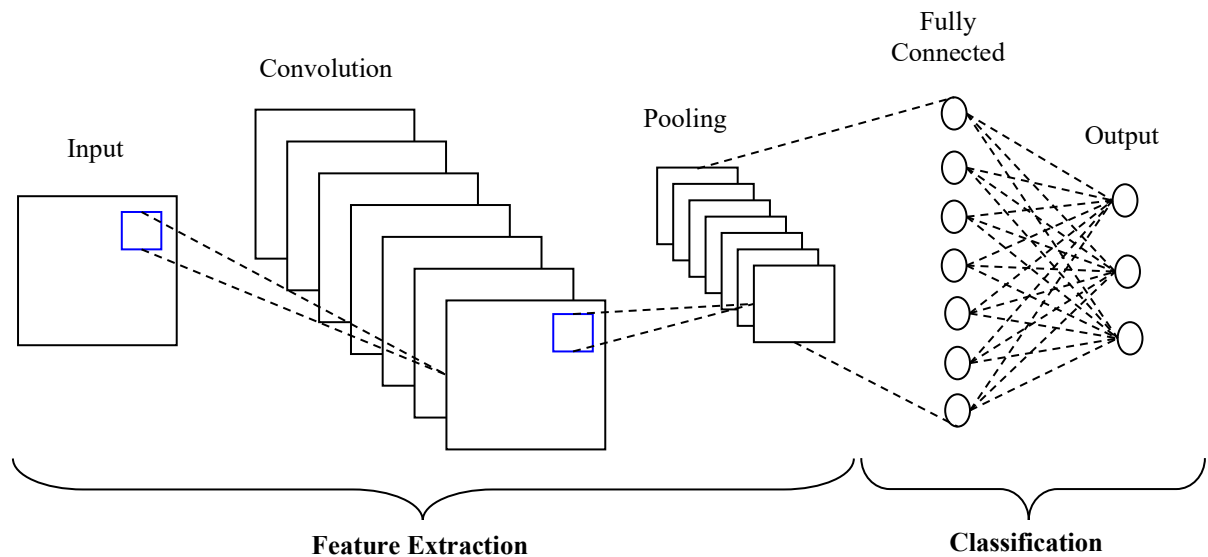

**Figure 5:** Architecture diagram of CNN

In this, the pooling layer is to perform the evaluation of maximum value from each segmented blocks of the image frame to form as a pre-feature matrix representation. This matrix is arranged as the sequential data to represent the feature vector for the input of convoluted image frame. This feature vector was processed in the fully-connected layer. This layer is act as the traditional process of neural network connectivity formation. From the output layer of neural network, the classified result of the image frame was presented as the index of matched vector which is related to the action of human from the video data.

4. RESULT ANALYSIS

The result analysis section evaluates the performance of proposed model of action recognition for the input video frame by comparing the statistical parameters of proposed work with other state-of-art methods. For this, UCF Sports video dataset and HMDB51 dataset are used to implement the proposed algorithm and testing the result by finding the maximum key points of image pattern. This proposed work is implemented and tested in the PYTHON version of V3.6 with required libraries. the validation of performance measures is calculated by finding the error difference between predicted result and ground truth of dataset. In this dataset, there are 51 discrete actions of classes are there. Among that, five number of actions are selected for the training and testing of human action from the video frames. This it contains 100 number of video dataset are used for analysis. In that, 70% are considered as the training and 30% are testing. There are three methods of existing systems were referred to compare the performance of proposed work which are can be listed as

1. Deep bidirectional long short-term memory (DBiLSTM) [41].
2. Deep Convolution Symmetric Neural Network with PCANET [42].
3. Weakly-supervised action localization (WSAL) [43].
4. Deep Convoluted Neural Network (DCNN) [44].

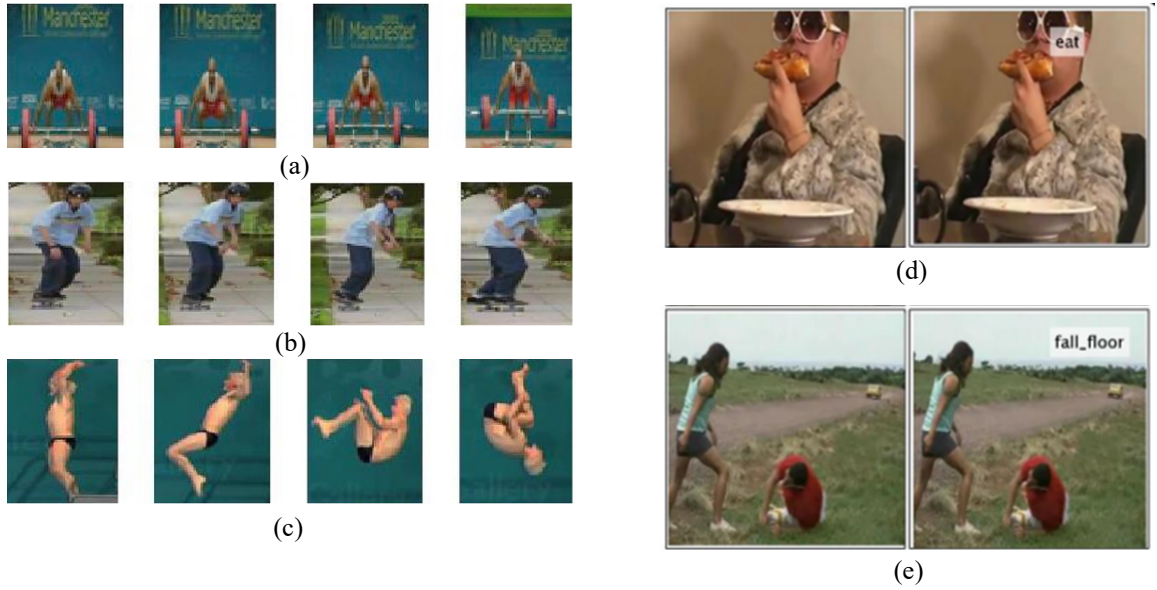

**Figure 6:** Sample video frames of UCF sports and HMDB51 dataset: (a) Lifting, (b) Skating, (c) Diving, (d) Eating, and (e) Fall on floor.

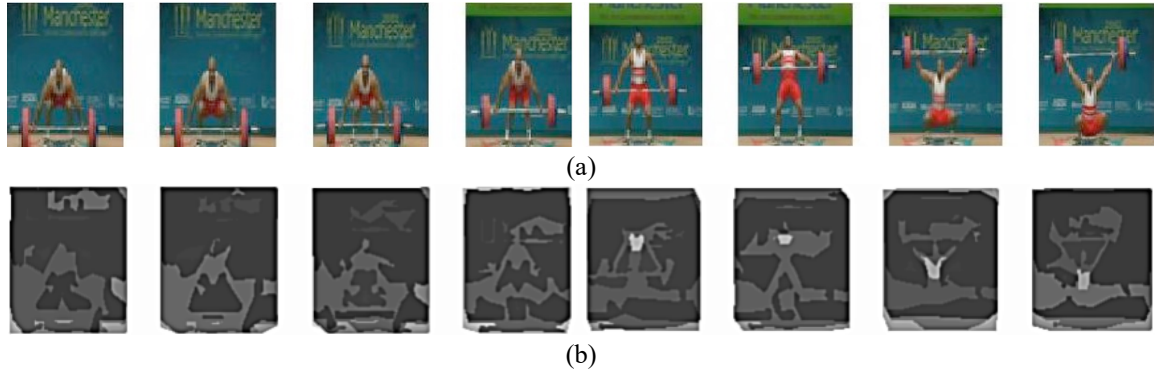

**Figure 7:** Sample video frames of UCF sports dataset: (a) Video frames, (b) Salient Features

The figure 6 shows the sample frames for the testing dataset. Figure 7 shows the Salient feature for weight lifting video frames from the CNN layer. The results can be compared with the existing methods by the statistical parameters such as, Sensitivity, Specificity, Precision, Recall, Jaccard, Dice Overlap, F1-Score, Matthew's correlation coefficient (MCC), Error rate, Kappa Coefficient and Accuracy which are all represented in the equations (14 to 24).

$$Sensitivity = \frac{TP}{TP + FN} \quad (14)$$

$$Specificity = \frac{TN}{TN + FP} \quad (15)$$

$$Jaccard\_Similarity = \frac{TP}{TP + FN + FP} \quad (16)$$

$$Dice\_Overlap = \frac{2TP}{FP + 2TP + FN} \quad (17)$$

$$Precision = \frac{TP}{TP + FP} \quad (18)$$

$$Recall = \frac{TP}{TP + FN} \quad (19)$$

$$F1\_Score = \frac{2 \times Precision \times Recall}{Precision + Recall} \quad (20)$$

$$MCC = \frac{TP \times TN - FP \times FN}{\sqrt{(TP+FP)(TP+FN)(TN+FP)(TN+FN)}} \quad (21)$$

$$Accuracy = \frac{TP+TN}{TP+TN+FP+FN} \quad (22)$$

$$Error\_Rate = 1 - Accuracy \quad (23)$$

$$Kappa\_Coeff = \frac{P_o - P_e}{1 - P_e} \quad (24)$$

Where, TP, FP, TN, and FN define the True Positive, False Positive, True Negative, and False Negative respectively. In the kappa coefficient, the relative observed agreement $P_o$ and hypothetical probability of chance agreement $P_e$ can be calculate by the equations (25) and (26).

$$P_o = \frac{TP+TN}{TP+TN+FP+FN} \quad (25)$$

$$P_e = \frac{((TP+FP)\times(TP+FN)) + ((TN+FP)\times(TN+FN))}{(TP+TN+FP+FN)^2} \quad (26)$$

The figure 8 and 9 shows the performance measure and the similarity rate for the clustering and classification of human action that compares the proposed work with the existing classification method of LSTM learning method. These graphs show the bar plot for Sensitivity, Specificity, Positive Predictive Value (PPV)and Accuracy by comparing the recognized result with ground-truth of the dataset. These graphs represent that the performance level of proposed algorithm was increased in the rage of ~98%. From this it shows that the error rate of proposed model is in the value of ~1%compare to other existing classification methods.

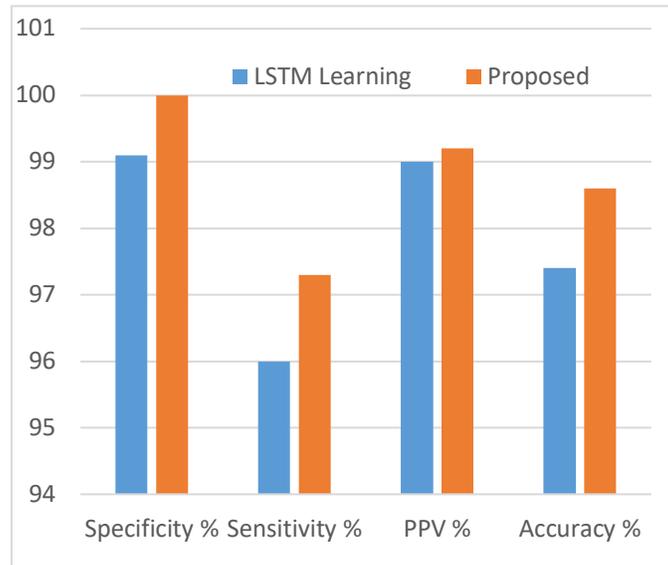

**Figure 8**: Performance Measures for UCF dataset

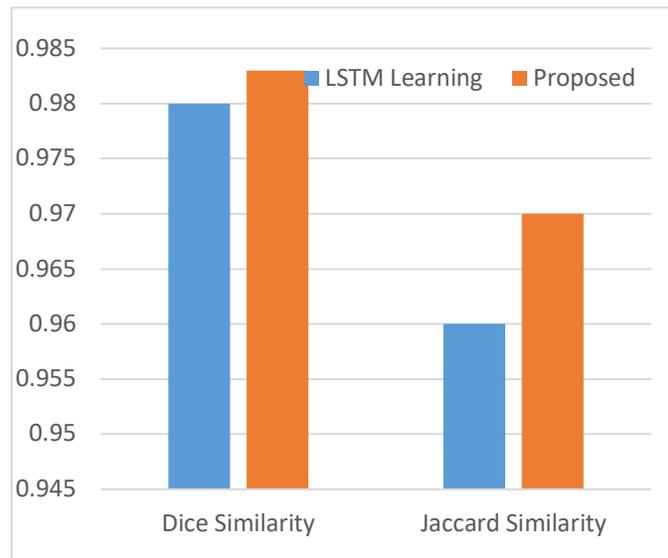
**Figure 9:** Similarity measure for UCF dataset

The figure 10 & 11 shows the comparison result of Accuracy, Kappa Coefficient and Receiver Operating Curve (ROC) for the ensemble pattern extraction method with existing method of [27] tested in UCF sports video dataset. The ROC curve is calculated from the confusion matrix and it is plotted for the False Positive Rate (FPR) and True Positive Rate (TPR). The FPR can be calculate by subtracting the Specificity from 1 and the True Positive Rate indicates the Sensitivity. This ROC reaches high sensitivity at minimum FPR which shows that the false rate of proposed work is less in the value of ~0.2.

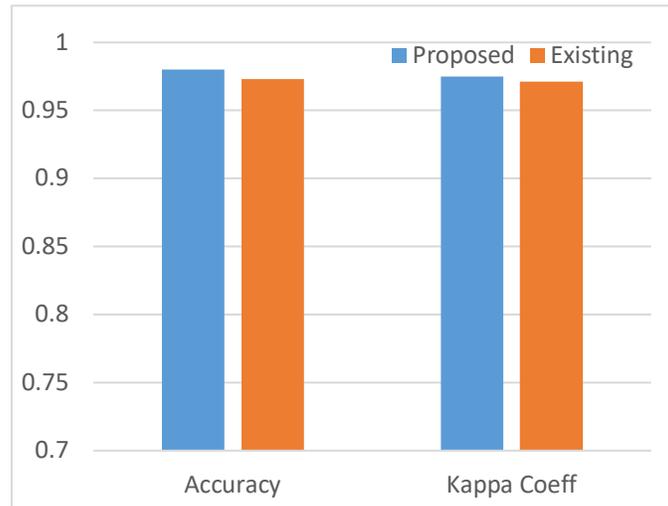
**Figure 10**: Accuracy chart for UCF dataset

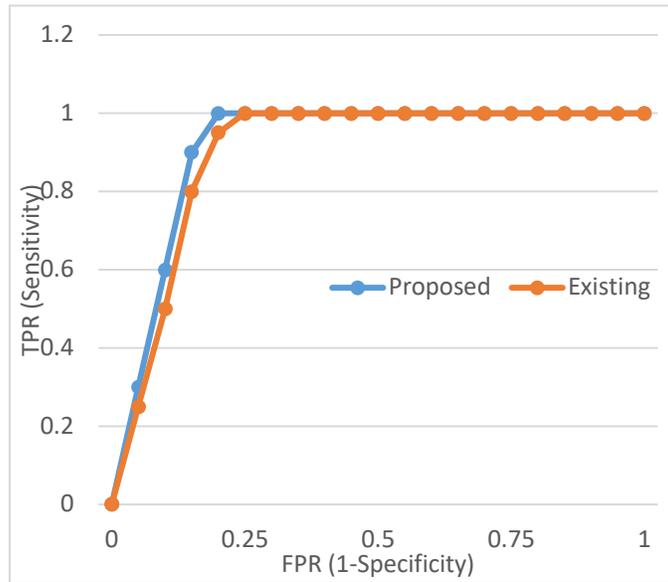

**Figure 11:** ROC Curve for UCF dataset

| | | 0 | 1 | 2 | 3 | 4 | 5 | 6 | 7 | 8 | 9 | 10 | 11 | 12 |
|---|---|---|---|---|---|---|---|---|---|---|---|---|---|---|
| | 0 | 360 | 0 | 0 | 0 | 1 | 0 | 1 | 0 | 2 | 0 | 1 | 3 | 0 |
| | 1 | 0 | 210 | 0 | 0 | 0 | 0 | 0 | 0 | 0 | 0 | 0 | 0 | 0 |
| | 2 | 0 | 0 | 301 | 0 | 0 | 0 | 0 | 0 | 0 | 0 | 0 | 0 | 0 |
| | 3 | 0 | 0 | 0 | 21 | 0 | 0 | 0 | 0 | 0 | 0 | 0 | 0 | 0 |
| | 4 | 0 | 0 | 0 | 0 | 417 | 0 | 0 | 0 | 0 | 0 | 0 | 0 | 0 |
| Actual | 5 | 0 | 0 | 0 | 0 | 0 | 97 | 0 | 0 | 0 | 0 | 0 | 0 | 0 |
| Class | 6 | 0 | 0 | 0 | 0 | 0 | 0 | 386 | 0 | 0 | 0 | 0 | 0 | 0 |
| Label | 7 | 0 | 0 | 0 | 0 | 0 | 0 | 0 | 256 | 0 | 0 | 0 | 0 | 0 |
| | 8 | 0 | 0 | 0 | 0 | 0 | 0 | 0 | 0 | 384 | 0 | 0 | 0 | 0 |
| | 9 | 0 | 0 | 0 | 0 | 0 | 0 | 0 | 0 | 0 | 251 | 1 | 0 | 2 |
| | 10 | 1 | 0 | 0 | 1 | 2 | 1 | 0 | 4 | 0 | 1 | 327 | 0 | 1 |
| | 11 | 0 | 1 | 0 | 1 | 0 | 0 | 2 | 1 | 2 | 0 | 0 | 279 | 0 |
| | 12 | 1 | 1 | 1 | 0 | 3 | 1 | 0 | 0 | 0 | 0 | 3 | 0 | 361 |
| | | 0 | 1 | 2 | 3 | 4 | 5 | 6 | 7 | 8 | 9 | 10 | 11 | 12 |
| | | Predicted Class Label | | | | | | | | | | | | |

**Figure 12** Confusion matrix for UCF dataset

Figure 12 shows the confusion matrix of the classified action for the overall video dataset. The confusion matrix is used to estimate the accuracy of the proposed work and to calculate the other parameters. In this, the diagonal of the matrix representing the correctly classified result and the remaining are the misclassified result.

**Table 1** Dice Score Comparison with [27] for UCF dataset

| Methods | Dice Score |
|---|---|
| S-BoF | 0.792 |
| T-BoF | 0.785 |
| ST-BoF | 0.883 |
| S-VLAD | 0.824 |
| T-VLAD | 0.796 |
| ST-VLAD | 0.886 |
| Proposed work | 0.983 |

**Table 2** PPV Comparison with [27] for UCF dataset

| Methods | PPV |
|---|---|
| S-BoF | 0.806 |
| T-BoF | 0.798 |
| ST-BoF | 0.871 |
| S-VLAD | 0.838 |
| T-VLAD | 0.812 |
| ST-VLAD | 0.902 |
| Proposed work | 0.992 |

**Table 3** Sensitivity Comparison with [27] for UCF dataset

| Methods | Sensitivity |
|---|---|
| S-BoF | 0.816 |
| T-BoF | 0.802 |
| ST-BoF | 0.865 |
| S-VLAD | 0.841 |
| T-VLAD | 0.823 |
| ST-VLAD | 0.917 |
| Proposed work | 0.973 |

**Table 4** Accuracy Comparison with [27] for UCF dataset

| Methods | Sensitivity |
|---|---|
| S-BoF | 0.82 |
| T-BoF | 0.813 |
| ST-BoF | 0.865 |
| S-VLAD | 0.844 |
| T-VLAD | 0.832 |
| ST-VLAD | 0.928 |
| Proposed work | 0.976 |

Table 1, 2, 3, and 4 represents the comparison table of proposed work with existing systems of [27] for the parameter of Dice Similarity, PPV, sensitivity and accuracy respectively. Also, the table 5 shows the comparison result of Quantitative results for different architecture implemented in UCF sports video dataset. The comparison for proposed work is with the WSAL method in [28]. This shows that the proposed model achieved better recognition rate due to the texture pattern analysis from video input.

**Table 5** Quantitative results of different architecture for UCF dataset

| Methods | Dice | Sensitivity | PPV | Acc |
|---|---|---|---|---|
| WRN | 0.806 | 0.815 | 0.823 | 0.833 |
| WRN-PPNet without original input | 0.811 | 0.817 | 0.819 | 0.824 |
| WSAL [28] | 0.823 | 0.818 | 0.827 | 0.829 |
| Proposed work | 0.983 | 0.973 | 0.992 | 0.976 |

The Dice similarity and sensitivity measurement for the methods of [A] Deep Neural Network, [B] CNN, [C] CNN with fully connected CRF, [D] WRN-PPNet, and [E] proposed method of classification is shown in the figure 13.

From the overall results, the performance of proposed work is justified with different methods of classification and feature representation. Also, the complexity of proposed work is discussed in the result discussion section.

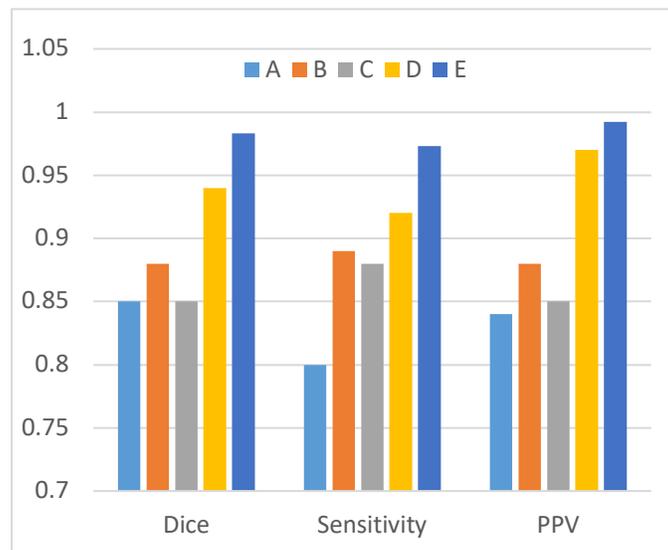

**Figure 13:** Quantitative results of the proposed and compared classification methods for UCF dataset

The table 6, 7, and 8 showing the comparison of proposed work with several existing model of CNN based on the performance measures such as Accuracy (%), Sensitivity and Specificity in (%) and Testing time (Sec). This was also represented in the figure 14 to 16. This was tested and analyzed for the HMDB51 dataset. In these table result, the accuracy of the proposed work is improved approximately 6.4% better than other CNN model in the [44] paper. Also, the time taken for the testing gets reduced which representing the reduction of time complexity in the proposed model of GWO with CNN classification.

**Table 6** Comparison table of Accuracy (%) for HMDB51 dataset

| Methods | Accuracy (%) | Error rate (%) |
|---|---|---|
| Proposed | 89.7 | 10.3 |
| ELM | 81.4 | 18.6 |
| Softmax | 74.8 | 25.2 |
| MSVM | 65.9 | 34.1 |
| KNN | 66.1 | 33.9 |
| Ensemble Tree | 63.4 | 36.6 |

**Table 7** Comparison table of Sensitivity (%) and Specificity (%) for HMDB51 dataset

| Methods | Sensitivity (%) | Specificity (%) |
|---|---|---|
| Proposed | 88.71 | 89.75 |
| ELM | 80.82 | 82.14 |
| Softmax | 74.19 | 74.83 |
| MSVM | 65.77 | 66.62 |
| KNN | 65.55 | 66.2 |
| Ensemble Tree | 62.91 | 63.93 |

**Table 7** Comparison table of training time and testing time (Sec) for HMDB51 dataset

| Methods | Training Time (Sec) | Testing Time (Sec) |
|---|---|---|
| Proposed | 411.243 | 176.247 |
| ELM | 493.514 | 211.506 |
| Softmax | 562.557 | 241.096 |
| MSVM | 920.694 | 394.583 |
| KNN | 725.704 | 311.016 |
| Ensemble Tree | 950.367 | 407.300 |

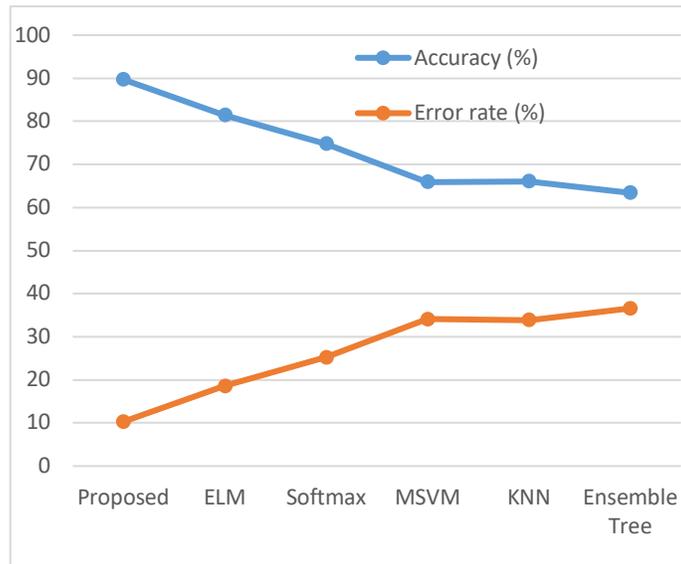

**Figure 14** Comparison chart of Accuracy (%) and Error Rate (%) for HMDB51 dataset

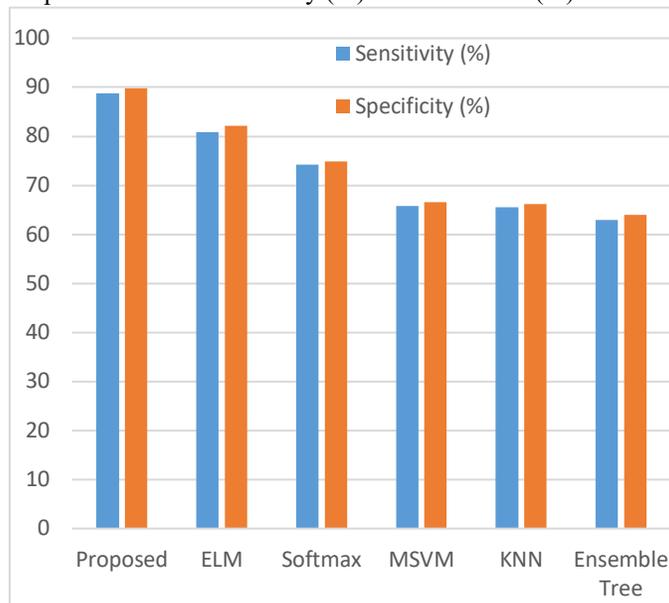

**Figure 15** Comparison chart of Sensitivity (%) and Specificity (%) for HMDB51 dataset

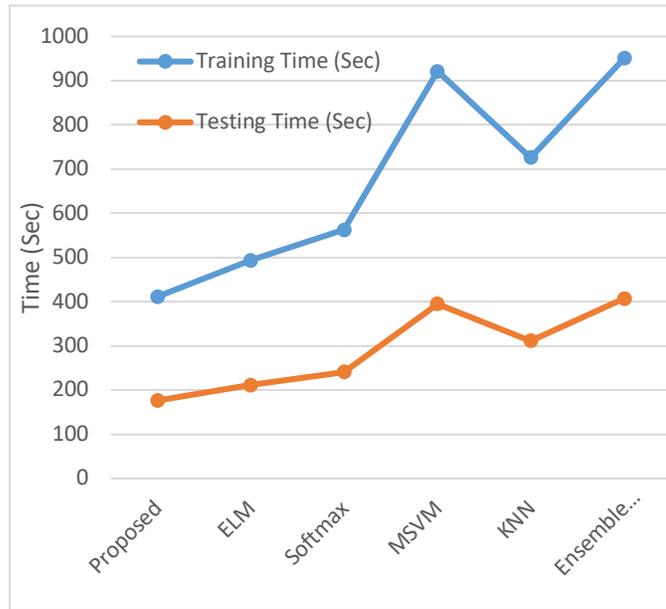

**Figure 16** Comparison chart of Training time (Sec) and Testing time (Sec) for HMDB51 dataset

## II. Discussion

In this section, it discusses about the reports of proposed work result for the application of action prediction for the video source input. The performance of the optimization process is depending on the complexity of the system. The complexity indicates time and space complexity of the algorithm that are consumed by the GWO method. The time complexity is depending on the number of iterations required to solve the problem and time taken for the one iteration.

In this model, the space complexity can be representing by the amount of feature space that are used to save as the feature database. In generally, the CNN requires the raw image to classify the image type. In the proposed work, instead of taking the raw image, the dimensionality reduced texture pattern was used for the classification step. By considering this, the space complexity is reduced compare to the traditional model of CNN classification. For the proposed model, the time complexity can be estimate and represent as in the form of notation $O\left(q \times \frac{\ln(p)}{h}\right)$. From the equation, the '$q$' indicated the total number of cycles taken for searching the best solution and logarithmic of time '$p$' for predicting the nearest match with the length of feature attributes '$h$'. From the overall work result, it reduces the complexity of model compare to other classification methods.

The result analysis justifies that the proposed model of texture pattern classification system achieved ~98% of recognition rate compares to the other traditional pattern analysis method.

## 5. Conclusion and Future enhancement

This paper mainly focused in the identification of object movement in image frame and to classify the action label of it. In that, the proposed feature extraction ensembles the spatial model of image pattern by using the image convolution method and the pattern model of feature extraction technique by the wavelet transform of the image frame. This type of feature extraction enhanced the pattern analysis model in the action prediction for video dataset. The CNN also makes the classification performance in feature analysis better than the other state-of-art methods. This was also enhanced by using the Grey Wolf Optimization method (GWO) by optimally selects the best attributes for neural network architecture arrangement.

In future, this type of optimal feature selection and classification can be implemented in the other image processing application to reduce the classification time complexity and also improve the performance furthermore. For the clustering and segmentation process, the iteration count can also be reducing by the optimal selection of cluster label which may further improve the performance rate.